\documentclass{bmvc2k}

\title{Random Word Data Augmentation with CLIP for Zero-Shot Anomaly Detection}

\addauthor{Masato Tamura}{masato.tamura@ieee.org}{1}

\addinstitution{
 Big Data Analytics Solutions Lab\\
 Hitachi America, Ltd.\\
 2535 Augustine Dr, Santa Clara,
 California USA
}

\runninghead{Masato Tamura}{Rand. Word Data Aug. for Zero-Shot Anomaly Detection}

\def\etal{\emph{et~al}\bmvaOneDot}

\usepackage{booktabs}
\usepackage{threeparttable}
\usepackage{multirow}
\usepackage{bm}
\usepackage{bbm}
\usepackage{siunitx}
\usepackage{amssymb}
\usepackage{arydshln}
\usepackage{nimbusmono}
\usepackage[T1]{fontenc}
\usepackage{wrapfig}

\begin{document}

\maketitle

\begin{abstract}
This paper presents a novel method that leverages a visual-language model, CLIP, as a data source for zero-shot anomaly detection. Tremendous efforts have been put towards developing anomaly detectors due to their potential industrial applications. Considering the difficulty in acquiring various anomalous samples for training, most existing methods train models with only normal samples and measure discrepancies from the distribution of normal samples during inference, which requires training a model for each object category. The problem of this inefficient training requirement has been tackled by designing a CLIP-based anomaly detector that applies prompt-guided classification to each part of an image in a sliding window manner. However, the method still suffers from the labor of careful prompt ensembling with known object categories. To overcome the issues above, we propose leveraging CLIP as a data source for training. Our method generates text embeddings with the text encoder in CLIP with typical prompts that include words of normal and anomaly. In addition to these words, we insert several randomly generated words into prompts, which enables the encoder to generate a diverse set of normal and anomalous samples. Using the generated embeddings as training data, a feed-forward neural network learns to extract features of normal and anomaly from CLIP's embeddings, and as a result, a category-agnostic anomaly detector can be obtained without any training images. Experimental results demonstrate that our method achieves state-of-the-art performance without laborious prompt ensembling in zero-shot setups.
\end{abstract}


\section{Introduction}\label{sec:intro}

\begin{figure}[tb]
  \centering
  \begin{minipage}[b]{1.0\linewidth}
    \centering
    \includegraphics[keepaspectratio,height=80pt]{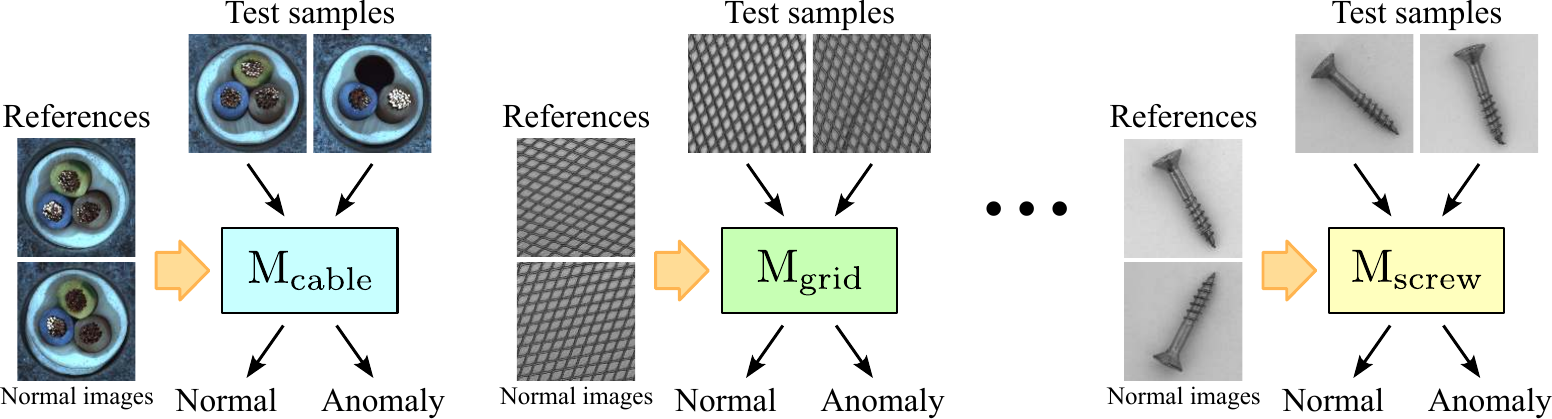}
    \subcaption{Vanilla AD.}\label{fig:intro_vanilla}
  \end{minipage} \\
  \begin{minipage}[b]{0.505\linewidth}
    \centering
    \includegraphics[keepaspectratio,height=80pt]{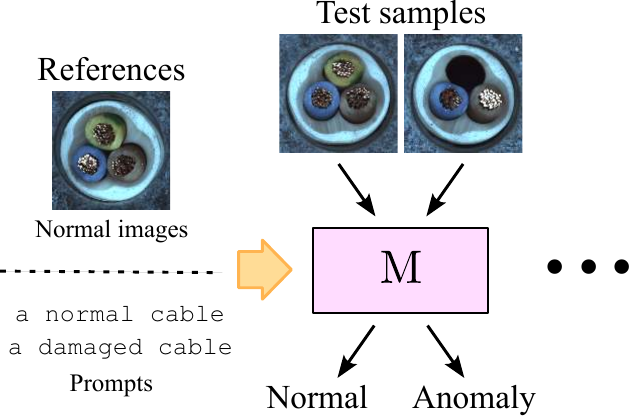}
    \subcaption{Category-agnostic known-object AD.}\label{fig:intro_catagno}
  \end{minipage}
  \hfill
  \begin{minipage}[b]{0.485\linewidth}
    \centering
    \includegraphics[keepaspectratio,height=80pt]{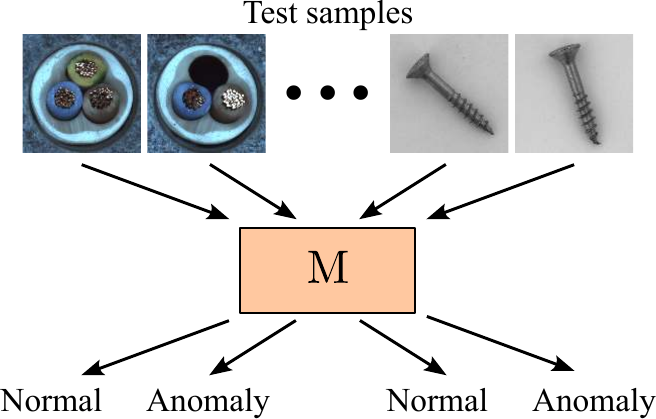}
    \subcaption{Category-agnostic unknown-object AD (ours).}\label{fig:intro_ours}
  \end{minipage}
  \vspace{-2em}
  \caption{Existing and our AD methods. (a) Existing vanilla AD methods train a model for each object category with normal samples as references and then test the models with the samples of the target object categories. (b) Existing category-agnostic methods do not require category-specific models. However, target object information such as normal images and prompts must be provided during inference. (c) Our method does not need either category-specific models or target object information during inference.}\label{fig:three graphs}
  \vspace{-1.0em}
\end{figure}

Visual anomaly detection (AD) is the task of classifying images as normal or anomaly, where anomalous images typically capture damaged, broken, or defective objects. AD differs from object classification in that anomalous samples are defined as those different from normal samples and thus are not restricted to specific categories. Due to the diverse appearances of anomalous samples, AD remains a challenging problem and thus has been tackled by numerous works~\cite{golan_nips2018,ruff_icml2018,gong_iccv2019,bergmann_cvpr2019,ye_ieeemm2019,yi_accv2020,cohen_arxiv2020,bergmann_cvpr2020,salehi_cvpr2020,rudolph_wacv2021,defard_icprw2021,zavrtanik_iccv2021,li_cvpr2021,sheynin_iccv2021,wu_iccv2021,roth_cvpr2022,zou_eccv2022,bergmann_ijcv2022,huang_eccv2022,ristea_cvpr2022,jeong_arxiv2023} for real-world applications.

Considering the fact that a diverse set of anomalous samples is difficult to be obtained, most existing methods~\cite{bergmann_cvpr2020,bergmann_ijcv2022,golan_nips2018,gong_iccv2019,li_cvpr2021,ristea_cvpr2022,rudolph_wacv2021,ruff_icml2018,salehi_cvpr2020,sheynin_iccv2021,wu_iccv2021,ye_ieeemm2019,yi_accv2020,zavrtanik_iccv2021,zou_eccv2022} train models with normal samples and then test on both normal and anomalous samples. The distribution of normal samples is modeled via various methods such as one-class classification~\cite{ruff_icml2018,yi_accv2020}, reconstruction~\cite{gong_iccv2019,wu_iccv2021}, and imitation of anomalous samples~\cite{li_cvpr2021,zavrtanik_iccv2021} during training, and discrepancies from the distribution are calculated to predict anomaly scores during inference. Since the methods are built on the assumption that normal samples are nearly identical, one model can only be applied to the objects of the same category as depicted in Fig.~\ref{fig:intro_vanilla}. This category-specific model requirement demands a significant number of models when the number of target objects is large, which renders the deployment of the methods impractical.

To overcome the aforementioned issue, category-agnostic methods~\cite{cohen_arxiv2020,defard_icprw2021,huang_eccv2022,jeong_arxiv2023,roth_cvpr2022} have been proposed. These methods leverage large-scale pre-trained models to extract highly generalizable features and utilize them as references to detect anomalous samples as illustrated in Fig.~\ref{fig:intro_catagno}. In particular, recently proposed WinCLIP~\cite{jeong_arxiv2023} demonstrates significant performance improvement over the other category-agnostic methods in a zero-shot AD setup, which extends a highly capable vison-language model, the Contrastive Language-Image Pre-training (CLIP) model~\cite{radford_icml2021}, to AD and is the most similar to our method.

In this paper, we also concentrate on zero-shot AD with CLIP. However, our method differs from WinCLIP in that we employ CLIP as a data source for training anomaly detectors. We leverage the observation that CLIP's text encoder generates perturbed output embeddings when input sentences are augmented with random words. By this augmentation, we produce a set of highly diverse embeddings with typical prompts containing words of normal and anomaly. These embeddings are then utilized as training data for a feed-forward neural network (FNN), which extracts normal and anomaly features from the embeddings and classifies them accordingly. Since FNN is trained without any object information, our proposed approach is applicable even when target objects are unknown during inference as shown in Fig.~\ref{fig:intro_ours}. Furthermore, our method does not require any laborious prompt ensembling, which is leveraged in WinCLIP to improve performance.

To summarize, our contributions are three-fold:
\begin{itemize}
  \item We propose a novel AD method that leverages CLIP as a data source for training an FNN. Since our method does not require any target object information during training or inference, the trained model can be applied to the case where anomalous samples of unknown objects must be detected.
  \vspace{-0.5em}
  \item Our method achieves competitive performance to state-of-the-art methods without any prompt ensembling on two benchmark datasets in challenging zero-shot setups.
  \vspace{-0.5em}
  \item Extensive experiments show the potential use case of the proposed method, where anomalous samples contain objects of ambiguous categories.
\end{itemize}

\section{Related Work}\label{sec:related}

\subsection{Anomaly Detection}\label{subsec:ac}

Most existing AD methods~\cite{bergmann_cvpr2020,bergmann_ijcv2022,golan_nips2018,gong_iccv2019,li_cvpr2021,ristea_cvpr2022,rudolph_wacv2021,ruff_icml2018,salehi_cvpr2020,sheynin_iccv2021,wu_iccv2021,ye_ieeemm2019,yi_accv2020,zavrtanik_iccv2021,zou_eccv2022} employ category-specific models to capture the distribution of normal samples during training and then calculate the discrepancies from the distribution during inference. 
One major trend of training category-specific models is to use unsupervised learning. An early attempt~\cite{ruff_icml2018} enhances support vector data
description, which is a classic algorithm of one-class classification, with deep neural networks. Yi and Yoon~\cite{yi_accv2020} extend the method to patch-level classification for precise anomaly segmentation. Gong~\etal~\cite{gong_iccv2019} employ another dominant unsupervised approach, where autoencoders are trained to reconstruct normal samples from anomalous ones.
Another trend of category-specific models is to leverage self-supervised learning. Golan and Yaniv~\cite{golan_nips2018} and Li~\etal~\cite{li_cvpr2021} synthesize anomalous samples from normal ones using image transformations and then train models to classify original and synthesized images. Ye~\etal~\cite{ye_ieeemm2019} and Riesta~\etal~\cite{ristea_cvpr2022} propose erasing information in images and forcing models to restore the information during training, which cannot restore anomalous samples perfectly during inference. The aforementioned methods achieve promising results with sophisticated ways of modeling the distribution. However, the inefficiency of category-specific models is non-negligible if the number of categories is significant and thus renders the methods impractical.

To solve the inefficiency of category-specific models, several category-agnostic methods~\cite{cohen_arxiv2020,defard_icprw2021,huang_eccv2022,jeong_arxiv2023,roth_cvpr2022} have been proposed, which leverage large-scale pre-trained models. Most of these methods~\cite {cohen_arxiv2020,defard_icprw2021,huang_eccv2022,roth_cvpr2022} extract features with ImageNet~\cite{deng_cvpr2009} pre-trained models to register the information of normal samples and utilize the information to find anomalous samples. Recently proposed WinCLIP~\cite{jeong_arxiv2023} differs from the ImageNet-based methods in that the method employs a vison-language model, CLIP~\cite{radford_icml2021}, for prompt-guided AD. All of these methods enable models to detect anomalous samples in either zero- and/or few-shot manner due to the generalization capability of the pre-trained models.

Our method also leverages CLIP but differs from WinCLIP in that we use CLIP as a data source for training anomaly detectors. Due to the category-agnostic and highly-diverse samples from CLIP, our method can be applied to samples of unknown object categories and achieve competitive performance without any prompt ensembling in zero-shot AD setups.

\subsection{Zero-Shot Image Recognition with CLIP}\label{subsec:zero_clip}

CLIP~\cite{radford_icml2021} is trained with a significant amount of image-text pairs for better generalization performance. During the training, embeddings from the image encoder are forced to have high cosine similarities with the text embeddings of the corresponding pairs from the text encoder. In contrast, the similarities of the embeddings in different pairs are reduced. This contrastive learning enables models to extract highly generalized features from images, and as a result, the models achieve high performance in zero-shot image classification.

Following the success of zero-shot image classification, several image recognition tasks have been tackled with CLIP in a zero-shot manner~\cite{gu_iclr2022,xu_eccv2022,sato_cvpr2023}. Gu~\etal~\cite{gu_iclr2022} and Xu~\etal~\cite{xu_eccv2022} extend the CLIP-based image classification to object detection and segmentation, respectively. Sato~\etal~\cite{sato_cvpr2023} tackles zero-shot anomaly action recognition with prompt-guided text embeddings for anomalous behaviors unobserved during training.

We also leverage CLIP but in a different manner; we use CLIP as a data source for training a classifier. The training samples generated by CLIP have high diversity and thus enhance the capability of anomaly detectors without prompt ensembling.

\section{Proposed Method}\label{sec:method}

We leverage CLIP~\cite{radford_icml2021} as a data source to detect anomaly samples in a zero-shot manner. To elaborate on our method, we first describe prompt-guided AD in Sec.~\ref{subsec:prompt} and then explain the proposed random word data augmentation in Sec.~\ref{subsec:randaug}, which is the core of our method. Finally, we illustrate the way to train an FNN with samples generated by CLIP in Sec.~\ref{subsec:traininf}. It should be noted that in this section, we explain AD in the case where target object categories are unknown.

\subsection{Prompt-Guided Anomaly Detection}\label{subsec:prompt}

Figure~\ref{fig:prompt} shows a prompt-guided AD method. We use two-class prompts as in WinCLIP~\cite{jeong_arxiv2023}. However, we do not use state or prompt ensembling, which is utilized in WinCLIP and requires laborious engineering. In prompt-guided AD, two-class prompts, ``{\renewcommand*\familydefault{\ttdefault}\normalfont a photo of [n] object}'' and ``{\renewcommand*\familydefault{\ttdefault}\normalfont a photo of [a] object}'', are first prepared, where words of normal and anomaly are inserted into the locations of ``{\renewcommand*\familydefault{\ttdefault}\normalfont [n]}'' and ``{\renewcommand*\familydefault{\ttdefault}\normalfont [a]}'', respectively. Examples of normal words are ``{\renewcommand*\familydefault{\ttdefault}\normalfont a}'' and ``{\renewcommand*\familydefault{\ttdefault}\normalfont a normal}'', and those of anomlay words are ``{\renewcommand*\familydefault{\ttdefault}\normalfont a damaged}'' and ``{\renewcommand*\familydefault{\ttdefault}\normalfont a broken}''. Two prompts are then transformed into tokens $\bm{t}^{(n)} \in \mathbb{Z}^{C_{t}}$ and $\bm{t}^{(a)} \in \mathbb{Z}^{C_{t}}$ with the tokenizer in CLIP, where $C_{t}$ is the maximum token size. The tokens are further transformed into text embeddings $\bm{e}^{(n,t)} \in \mathbb{R}^{C_{e}}$ and $\bm{e}^{(a,t)} \in \mathbb{R}^{C_{e}}$ as $\bm{e}^{(n,t)} = f_{tenc}(\bm{t}^{(n)})$ and $\bm{e}^{(a,t)} = f_{tenc}(\bm{t}^{(a)})$, where $C_{e}$ is the embedding dimension and $f_{tenc}(\cdot)$ is the text encoder in CLIP. The obtained two embeddings are used as guides for AD.

To calculate anomaly scores with the text embeddings, input images must be transformed into embeddings. Given an input image $I \in \mathbb{R}^{3\times H\times W}$, where $H$ and $W$ are the height and width of the input image, an image embedding $\bm{e}^{(i)} \in \mathbb{R}^{C_{e}}$ is obtained as $\bm{e}^{(i)} = f_{ienc}(I)$, where $f_{ienc}(\cdot)$ is the image encoder in CLIP. After normalizing the embeddings as $\bar{\bm{e}}^{(n,t)} = \frac{\bm{e}^{(n,t)}}{|\bm{e}^{(n,t)}|}$, $\bar{\bm{e}}^{(a,t)} = \frac{\bm{e}^{(a,t)}}{|\bm{e}^{(a,t)}|}$, and $\bar{\bm{e}}^{(i)} = \frac{\bm{e}^{(i)}}{|\bm{e}^{(i)}|}$, an anomaly score $s_{pr} \in [0, 1]$ is finally obtained using the softmax as $s_{pr} = \frac{\exp(\bar{\bm{e}}^{(a,t)}\cdot \bar{\bm{e}}^{(i)} / T)}{\exp(\bar{\bm{e}}^{(n,t)}\cdot \bar{\bm{e}}^{(i)} / T) + \exp(\bar{\bm{e}}^{(a,t)}\cdot \bar{\bm{e}}^{(i)} / T)}$, where $T$ is a temperature parameter to adjust the sensitivity of the softmax. Following the implementation of CLIP~\cite{radford_icml2021}, we set the temperature parameter $T$ to 0.01.

Since the prompts do not specify any object categories, this prompt-guided AD can be applied when object categories are unknown. The method can be modified for the known-object case by changing the word ``{\renewcommand*\familydefault{\ttdefault}\normalfont object}'' in the prompts to a target category name.

\subsection{Random Word Data Augmentation}\label{subsec:randaug}

\begin{figure}[tb]
  \centering
  \begin{minipage}[b]{0.35\linewidth}
    \centering
    \includegraphics[keepaspectratio,height=79pt]{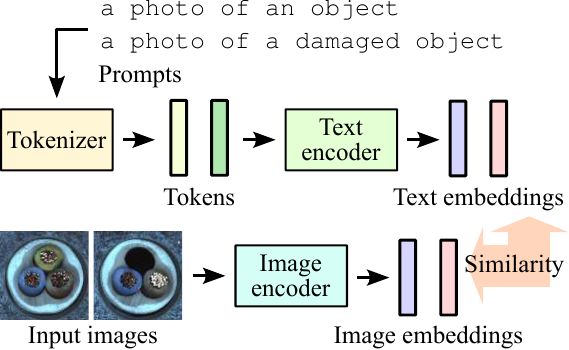}
    \subcaption{Prompt-guided AD.}\label{fig:prompt}
  \end{minipage}
  \hfill
  \begin{minipage}[b]{0.64\linewidth}
    \centering
    \includegraphics[keepaspectratio,height=79pt]{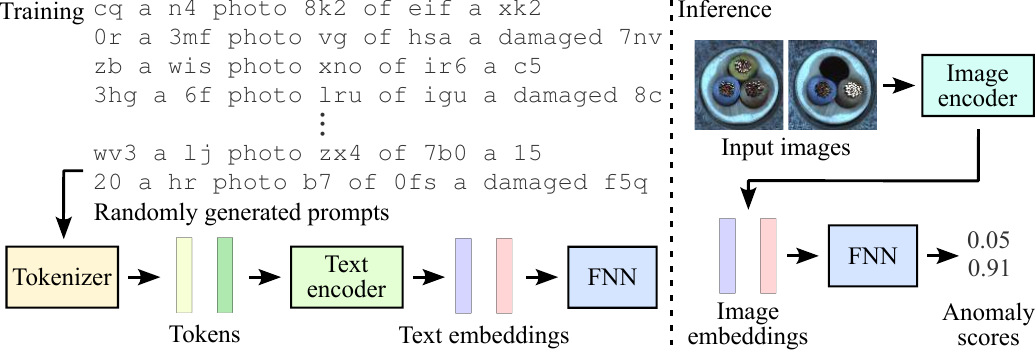}
    \subcaption{Our method.}\label{fig:ours}
  \end{minipage}
  \vspace{-2em}
  \caption{Overview of prompt-guided AD and our method.}\label{fig:overview}
  \vspace{-1em}
\end{figure}

\begin{wrapfigure}{r}{0.35\textwidth}
  \vspace{-2em}
  \begin{center}
    \includegraphics[width=0.35\textwidth]{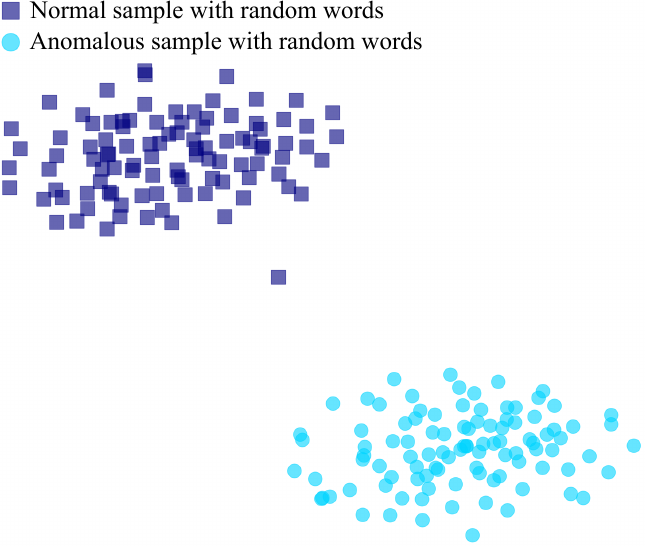}
  \end{center}
  \vspace{-2em}
  \caption{t-SNE plot of generated text embeddings.}\label{fig:trainfeat}
  \vspace{-1em}
\end{wrapfigure}

To train an FNN-based anomaly detector, we generate training samples by feeding prompts into CLIP with random word data augmentation. Samples are generated based on two prompt templates, ``{\renewcommand*\familydefault{\ttdefault}\normalfont [$\text{w}_{\text{0}}$] a [$\text{w}_{\text{1}}$] photo [$\text{w}_{\text{2}}$] of 
[$\text{w}_{\text{3}}$] [n] [$\text{w}_{\text{4}}$]}'' for normal samples and ``{\renewcommand*\familydefault{\ttdefault}\normalfont [$\text{w}_{\text{5}}$] a [$\text{w}_{\text{6}}$] photo [$\text{w}_{\text{7}}$] of 
[$\text{w}_{\text{8}}$] [a] [$\text{w}_{\text{9}}$]}'' for anomalous samples. At the locations of ``{\renewcommand*\familydefault{\ttdefault}\normalfont [n]}'' and ``{\renewcommand*\familydefault{\ttdefault}\normalfont [a]}'', words of normal and anomaly are inserted, respectively, in the same way as the prompt-guided AD. At the locations of ``{\renewcommand*\familydefault{\ttdefault}\normalfont [$\text{w}_{\text{i}}$]}'', randomly generated words are inserted.

Words for ``{\renewcommand*\familydefault{\ttdefault}\normalfont [$\text{w}_{\text{i}}$]}'' are generated by randomly selecting letters from the English alphabet and digits. The length of each word is also randomly selected between the range from the minimum length $l_{min}$ to the maximum length $l_{max}$. The selected letters are concatenated and compose a single word. We insert individually generated words into the prompt templates and finally obtain prompts as depicted in Fig.~\ref{fig:ours} for generating training samples.

The completed prompts are transformed into text embeddings via the tokenizer and text encoder in CLIP. Suppose we have $N_{p}$ pairs of completed normal and anomalous prompts, a set of the pairs of tokens $\mathcal{T} = \{(\bm{t}_{i}^{(n)}, \bm{t}_{i}^{(a)})\}_{i=1}^{N_{p}}$ is obtained, and then tokens $\bm{t}_{i}^{(n)}$ and $\bm{t}_{i}^{(a)}$ are transformed into text embeddings as $\bm{e}_{i}^{(n)} = f_{tenc}(\bm{t}_{i}^{(n)})$ and $\bm{e}_{i}^{(a)} = f_{tenc}(\bm{t}_{i}^{(a)})$. The obtained set of text embedding pairs $\mathcal{E} = \{(\bm{e}_{i}^{(n)}, \bm{e}_{i}^{(a)})\}_{i=1}^{N_{p}}$ is used as a training set for anomaly detectors. We evaluate AD performances with several $N_{p}$ values, which are reported in Sec.~\ref{subsec:anada}.

The proposed data augmentation increases the diversity of training samples and thus enhances the performance of anomaly detectors. Figure~\ref{fig:trainfeat} illustrates the t-SNE~\cite{maaten_jmlr2008} plot of the generated text embeddings. It can be seen from the figure that normal and anomalous samples are separately distributed, while samples in either group of normal or anomaly have a certain amount of diversity. Since any object categories are specified in the prompts to generate training samples, the trained anomaly detector can be applied even when target object categories are unknown, resulting in obtaining category-agnostic anomaly detectors.

\subsection{Training and Inference}\label{subsec:traininf}

As illustrated in Fig.~\ref{fig:ours}, we use an FNN as an anomaly detector, which is trained with the generated text embeddings. During training, generated samples $\bm{e}_{i}^{(n)}$ and $\bm{e}_{i}^{(a)}$ are fed into an FNN, and the binary cross entropy loss is calculated, where normal and anomalous samples are labeled as 0 and 1, respectively. The samples of the same pair are always contained in the same batch because we empirically found that the performance was slightly improved with that strategy. During inference, an input image $I$ is first transformed into an image embedding $\bm{e}^{(i)}$ as $\bm{e}^{(i)} = f_{ienc}(I)$, and then an anomaly score $s_{FNN}$ is obtained as $s_{FNN} = \sigma(f_{FNN}(\bm{e}^{(i)}))$, where $f_{FNN}(\cdot)$ is the trained FNN and $\sigma(\cdot)$ is the sigmoid function. Since the text and image embeddings from CLIP have similar semantics due to the contrastive learning of CLIP, the trained FNN can be applied to the image embeddings even though the FNN is not trained with them. The obtained scores can solely be used to detect anomalous samples or can be combined with any kind of score such as $s_{pr}$ to enhance the performance.

\section{Experiments}\label{sec:exp}

\subsection{Datasets and Evaluation Metrics}

To validate the effectiveness of the proposed method, we conduct extensive experiments on three publicly available benchmark datasets: the MVTec-AD~\cite{bergmann_cvpr2019}, VisA~\cite{zou_eccv2022}, and SewerML~\cite{haurum_cvpr2021} datasets. The former two datasets are common AD datasets. The MVTec-AD dataset includes 15 object categories, while the VisA dataset comprises 12 object categories. We use the SewerML dataset to evaluate the capability of detecting anomalous samples whose object categories are difficult to define due to their diversities. Although defect types are defined in the dataset, the appearance of each defect is not fixed and thus is challenging to classify, especially in a zero-shot manner. We compare CLIP~\cite{radford_icml2021} and our method with this dataset to validate the effectiveness when ambiguous anomalous samples must be detected. Only the validation set of the SewerML dataset is used for this evaluation.

Following the work of WinCLIP~\cite{jeong_arxiv2023}, we report the evaluation metrics of Area Under the Receiver Operating Characteristic (AUROC), Area Under the Precision-Recall curve (AUPR), and $F_{\text{1}}$-score at optimal threshold ($F_{\text{1}}$-max). The reported values of our method are the mean and standard deviation of trials with 10 random seeds.

\subsection{Implementation Details}

In all the experiments, we use OpenCLIP\footnote{\url{https://github.com/mlfoundations/open_clip}} as the implementation of CLIP~\cite{radford_icml2021} and the LAION-400M~\cite{schuhmann_arxiv2021}-based CLIP with ViT-B/16+~\cite{gabriel_soft2021}, which is also used by the work of WinCLIP~\cite{jeong_arxiv2023}, for fair comparisons. The FNN has four linear layers with batch normalization~\cite{ioffe_icml2015}, ReLU activations, and dropout~\cite{srivastava_jmlr2014}. The network is trained for 2 epochs using the AdamW~\cite{loshchilov_iclr2018} optimizer with a batch size of 128, the initial learning rate of $10^{-3}$ and the weight decay of $10^{-4}$. The learning rate is decayed after 1 epoch. Note that each batch comprises paired normal and anomalous samples as described in Sec.~\ref{subsec:traininf}. Unless otherwise noted, we use multi-crop data augmentation at test time for both CLIP and our method because WinCLIP also uses the multi-crop strategy.

For the random word data augmentation, the minimum word length $l_{min}$ is set to 5, and the maximum word length $l_{max}$ is set to 10. By default, 10,000 pairs of normal and anomalous samples are generated with ``{\renewcommand*\familydefault{\ttdefault}\normalfont a}'' for a word of normal and ``{\renewcommand*\familydefault{\ttdefault}\normalfont a damaged}'' for a word of anomaly. The performances with other settings are analyzed in Sec.~\ref{subsec:anada}.

As denoted in Sec.~\ref{subsec:traininf}, the scores of the FNN can be combined with other scores. In zero-shot setups, the performances with $s_{pr} + s_{FNN}$ (indicated by ``CLIP + ours'') as well as with the individual score $s_{FNN}$ (indicated by ``ours'') are reported. For future reference, we also report the performances of few-shot setups. In these setups, an additional score $s_{img}$ is calculated based on the similarities between the image embeddings of test samples and reference normal samples as described in the paper of WinCLIP. The performances are evaluated with the added score $s_{pr} + s_{FNN} + s_{img}$ (indicated by ``CLIP + ours'').

\subsection{Comparison against State of The Art}

\begin{table}[t]
  \centering
  \caption{Comparison against state-of-the-art methods.}\label{tab:comp}
  \footnotesize
  \begin{tabular}{@{}clcccccc@{}}
    \toprule
    && \multicolumn{3}{c}{MVTec-AD} & \multicolumn{3}{c}{VisA} \\
    \cmidrule(lr){3-5}\cmidrule(lr){6-8}
    Setup & Method & AUROC & AUPR & $F_{1}$-max & AUROC & AUPR & $F_{1}$-max \\
    \midrule
    \multirow{3}{*}{\shortstack{0-shot\\ $\left(\begin{array}{@{}c@{}}\text{Object}\\\text{unknown}\end{array}\right)$}} & CLIP~\cite{radford_icml2021} & 91.5 & 95.7 & 92.0 & 76.5 & 80.5 & 78.1 \\ [0.5ex]\cdashline{2-8}\noalign{\vskip 0.5ex}
    & Ours & 91.0$\pm$0.4 & 95.4$\pm$0.3 & 92.2$\pm$0.3 & 78.1$\pm$1.1 & 81.3$\pm$0.9 & 79.8$\pm$0.4 \\
    & CLIP + ours & \textbf{92.2$\pm$0.3} & \textbf{96.0$\pm$0.2} & \textbf{92.8$\pm$0.2} & \textbf{78.2$\pm$0.8} & \textbf{81.5$\pm$0.8} & \textbf{79.9$\pm$0.3} \\
    \midrule
    \multirow{3}{*}{\shortstack{0-shot\\ $\left(\begin{array}{@{}c@{}}\text{Object}\\\text{known}\end{array}\right)$}} & WinCLIP~\cite{jeong_arxiv2023} & 91.8 & \textbf{96.5} & 92.9 & 78.1 & 81.2 & 79.0 \\
    & CLIP~\cite{radford_icml2021} & 92.6 & 96.3 & 93.0 & 76.3 & 80.4 & 78.8 \\ [0.5ex]\cdashline{2-8}\noalign{\vskip 0.5ex}
    & CLIP + ours & \textbf{93.0$\pm$0.3} & 96.4$\pm$0.2 & \textbf{93.1$\pm$0.2} & \textbf{79.8$\pm$0.6} & \textbf{82.8$\pm$0.6} & \textbf{79.9$\pm$0.2} \\
    \midrule
    \multirow{5}{*}{1-shot} & SPADE~\cite{cohen_arxiv2020} & 81.0$\pm$2.0 & 90.6$\pm$0.8 & 90.3$\pm$0.8 & 79.5$\pm$4.0 & 82.0$\pm$3.3 & 80.7$\pm$1.9 \\
    & PaDiM~\cite{defard_icprw2021} & 76.6$\pm$3.1 & 88.1$\pm$1.7 & 88.2$\pm$1.1 & 62.8$\pm$5.4 & 68.3$\pm$4.0 & 75.3$\pm$1.2 \\
    & PatchCore~\cite{roth_cvpr2022} & 83.4$\pm$3.0 & 92.2$\pm$1.5 & 90.5$\pm$1.5 & 79.9$\pm$2.9 & 82.8$\pm$2.3 & 81.7$\pm$1.6 \\
    & WinCLIP~\cite{jeong_arxiv2023} & 93.1$\pm$2.0 & 96.5$\pm$0.9 & 93.7$\pm$1.1 & \textbf{83.8$\pm$4.0} & 85.1$\pm$4.0 & 83.1$\pm$1.7 \\ [0.5ex]\cdashline{2-8}\noalign{\vskip 0.5ex}
    & CLIP + ours & \textbf{93.3$\pm$0.5} & \textbf{96.7$\pm$0.2} & \textbf{94.0$\pm$0.3} & 83.4$\pm$1.7 & \textbf{85.8$\pm$1.8} & \textbf{83.6$\pm$0.8} \\
    \midrule
    \multirow{6}{*}{2-shot} & SPADE~\cite{cohen_arxiv2020} & 82.9$\pm$2.6 & 91.7$\pm$1.2 & 91.1$\pm$1.0 & 80.7$\pm$5.0 & 82.3$\pm$4.3 & 81.7$\pm$2.5 \\
    & PaDiM~\cite{defard_icprw2021} & 78.9$\pm$3.1 & 89.3$\pm$1.7 & 89.2$\pm$1.1 & 67.4$\pm$5.1 & 71.6$\pm$3.8 & 75.7$\pm$1.8 \\
    & PatchCore~\cite{roth_cvpr2022} & 86.3$\pm$3.3 & 93.8$\pm$1.7 & 92.0$\pm$1.5 & 81.6$\pm$4.0 & 84.8$\pm$3.2 & 82.5$\pm$1.8 \\
    & RegAD~\cite{huang_eccv2022} & 85.7 & -- & -- & -- & -- & -- \\
    & WinCLIP~\cite{jeong_arxiv2023} & \textbf{94.4$\pm$1.3} & \textbf{97.0$\pm$0.7} & \textbf{94.4$\pm$0.8} & 84.6$\pm$2.4 & 85.8$\pm$2.7 & 83.0$\pm$1.4 \\ [0.5ex]\cdashline{2-8}\noalign{\vskip 0.5ex}
    & CLIP + ours & 94.0$\pm$0.7 & 96.9$\pm$0.3 & 94.1$\pm$0.3 & \textbf{85.6$\pm$1.4} & \textbf{87.5$\pm$1.6} & \textbf{84.1$\pm$1.0} \\
    \midrule
    \multirow{6}{*}{4-shot} & SPADE~\cite{cohen_arxiv2020} & 84.8$\pm$2.5 & 92.5$\pm$1.2 & 91.5$\pm$0.9 & 81.7$\pm$3.4 & 83.4$\pm$2.7 & 82.1$\pm$2.1 \\
    & PaDiM~\cite{defard_icprw2021} & 80.4$\pm$2.5 & 90.5$\pm$1.6 & 90.2$\pm$1.2 & 72.8$\pm$2.9 & 75.6$\pm$2.2 & 78.0$\pm$1.2 \\
    & PatchCore~\cite{roth_cvpr2022} & 88.8$\pm$2.6 & 94.5$\pm$1.5 & 92.6$\pm$1.6 & 85.3$\pm$2.1 & 87.5$\pm$2.1 & 84.3$\pm$1.3 \\
    & RegAD~\cite{huang_eccv2022} & 88.2 & -- & -- & -- & -- & -- \\
    & WinCLIP~\cite{jeong_arxiv2023} & \textbf{95.2$\pm$1.3} & \textbf{97.3$\pm$0.6} & \textbf{94.7$\pm$0.8} & \textbf{87.3$\pm$1.8} & \textbf{88.8$\pm$1.8} & 84.2$\pm$1.6 \\ [0.5ex]\cdashline{2-8}\noalign{\vskip 0.5ex}
    & CLIP + ours & 94.5$\pm$0.7 & 97.1$\pm$0.3 & 94.4$\pm$0.3 & 86.6$\pm$0.9 & 88.4$\pm$1.3 & \textbf{84.5$\pm$0.6} \\
    \bottomrule
  \end{tabular}  
  \vspace{-1.0em}
\end{table}

We compare our method against state-of-the-art zero/few-shot AD methods. Table~\ref{tab:comp} shows the comparison results. For zero-shot AD, we evaluate performances in two setups. One is an unknown-object setup, where we assume that target object categories are unknown and thus a word ``{\renewcommand*\familydefault{\ttdefault}\normalfont object}'' is used in CLIP's prompts. The other is known-object setup, where we assume that target object categories are known and thus inserted into CLIP's prompts. As can be seen in the table, our method outperforms CLIP's~\cite{radford_icml2021} prompt-guided AD and WinCLIP~\cite{jeong_arxiv2023} with almost all the evaluation metrics when combined with CLIP. In particular, the performance gaps between the existing and our methods are larger in the zero-shot unknown-object setup than in the known-object setup, which indicates the effectiveness of detecting ambiguous anomalous samples.

The performances of the few-shot setups show that our method achieves competitive performances even when some normal reference images are provided. These results suggest that our method can be combined with state-of-the-art methods and enhance their performance.

\begin{figure}[t]
  \begin{minipage}[b]{0.56\textwidth}
    \centering
    \captionof{table}{Comparison against state-of-the-art methods without multiple crops on the MVTec-AD dataset in the zero-shot known-object setup.}\label{tab:compnocrop}
    \footnotesize
    \setlength{\tabcolsep}{4pt}
    \begin{tabular}{@{}lcccc@{}}
      \toprule
      Method & Prompt ens. & AUROC & AUPR & $F_{1}$-max \\
      \midrule
      WinCLIP~\cite{jeong_arxiv2023} & \checkmark & 90.8 & 96.1 & \textbf{92.5} \\
      CLIP~\cite{radford_icml2021} & & 89.8 & 95.4 & 92.1 \\ [0.5ex]\hdashline\noalign{\vskip 0.5ex}
      Ours & & 89.6$\pm$0.6 & 95.5$\pm$0.2 & 91.5$\pm$0.3 \\
      CLIP + ours & & \textbf{91.0$\pm$0.3} & \textbf{96.2$\pm$0.2} & \textbf{92.5$\pm$0.2} \\
      \bottomrule
    \end{tabular}
    \captionof*{table}{} 
  \end{minipage}
  \hfill
  \begin{minipage}[b]{0.42\textwidth}
    \centering
    \includegraphics[width=1.0\textwidth]{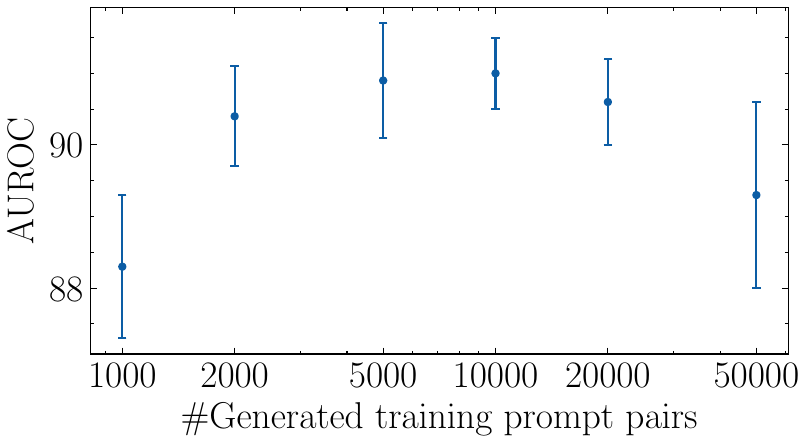}
    \vspace{-2em}
    \caption{AUROCs with the various numbers of training prompt pairs in the zero-shot unknown-object setup.}\label{fig:numwds}
  \end{minipage}
  \vspace{-1em}
\end{figure}

We also compare the performances of CLIP~\cite{radford_icml2021}, WinCLIP~\cite{jeong_arxiv2023}, and our method without the multi-crop strategy. If the multi-crop is eliminated from WinCLIP, the performance purely depends on the prompt ensembling, which is a suitable baseline to evaluate the capability of our method. Table~\ref{tab:compnocrop} shows the comparison results. As shown in the table, our method achieves better performance than WinCLIP. The results indicate that our random word data augmentation can improve performance better than prompt ensembling.

\subsection{Analysis on Data Augmention}~\label{subsec:anada}

We first analyze the effect of the number of prompt pairs $N_{p}$ for training the FNN. Figure~\ref{fig:numwds} shows the performances on the MVTec-AD dataset with regard to the different values of $N_{p}$ in the unknown-object setup. As shown in the figure, the best performance is achieved when $N_{p}$ is 10,000, and fewer or more pairs degrade the performance. The lower performance with fewer samples is attributed to insufficient training data, and that with more samples is probably due to over-fitting. As analyzed in Sec.~\ref{subsec:anafe}, embeddings generated from our random data augmentation have a different distribution from that of the embeddings generated from natural sentences. A large number of such training samples force models to have an incorrect decision boundary and thus degrades the performance.

\begin{table}[tb]
  \centering
  \caption{AUROCs with various word pairs for normal and anomaly in the zero-shot unknown-object setup. The left value in each cell is the result of CLIP~\cite{radford_icml2021}, and the right value is that of CLIP + ours.}\label{tab:wpair}
  \footnotesize
  \begin{tabular}{@{}lcccc@{}}
    \toprule
    & ``{\renewcommand*\familydefault{\ttdefault}\normalfont a damaged}'' & ``{\renewcommand*\familydefault{\ttdefault}\normalfont a broken}'' & ``{\renewcommand*\familydefault{\ttdefault}\normalfont a defective}'' & ``{\renewcommand*\familydefault{\ttdefault}\normalfont an anomalous}'' \\
    \midrule
    ``{\renewcommand*\familydefault{\ttdefault}\normalfont an}'' & 91.5$\diagup$\textbf{92.2$\pm$0.3} & \textbf{87.5}$\diagup$86.1$\pm$0.6 & 79.4$\diagup$\textbf{85.7$\pm$0.5} & 67.6$\diagup$\textbf{73.7$\pm$0.3} \\
    ``{\renewcommand*\familydefault{\ttdefault}\normalfont a normal}'' & 89.3$\diagup$\textbf{90.5$\pm$0.2} & 87.3$\diagup$\textbf{88.6$\pm$0.5} & 81.8$\diagup$\textbf{84.5$\pm$0.9} & 69.1$\diagup$\textbf{71.9$\pm$1.0} \\
    ``{\renewcommand*\familydefault{\ttdefault}\normalfont a good}'' & 88.4$\diagup$\textbf{89.6$\pm$0.3} & 86.1$\diagup$\textbf{87.0$\pm$0.5} & 80.6$\diagup$\textbf{86.3$\pm$0.4} & 68.6$\diagup$\textbf{73.0$\pm$0.9} \\
    ``{\renewcommand*\familydefault{\ttdefault}\normalfont a flawless}'' & 88.5$\diagup$\textbf{90.3$\pm$0.4} & 85.7$\diagup$\textbf{86.0$\pm$0.8} & 77.7$\diagup$\textbf{84.5$\pm$0.6} & 68.8$\diagup$\textbf{75.8$\pm$0.5} \\
    \bottomrule
  \end{tabular}
  \vspace{-2em}
\end{table}

We then analyze the effect of normal and anomaly word choices in guiding prompts and prompt templates. Table~\ref{tab:wpair} shows the evaluation results of CLIP$\diagup$CLIP + ours. Each row and column shows the performances with the indicated word as normal and anomaly in the prompts. As illustrated in the table, our method improves the performances of CLIP with any word pairs except for the pair of ``{\renewcommand*\familydefault{\ttdefault}\normalfont an}'' and ``{\renewcommand*\familydefault{\ttdefault}\normalfont a broken}''. In particular, our method significantly improves performances in the case where word pairs are inappropriate for CLIP. The results suggest that our method can show word-agnostic performance improvement and can boost the robustness of the prompt-guided AD.

\subsection{Analysis on Feature Embeddings}\label{subsec:anafe}

\begin{figure}[tb]
  \centering
  \includegraphics[width=1.0\linewidth]{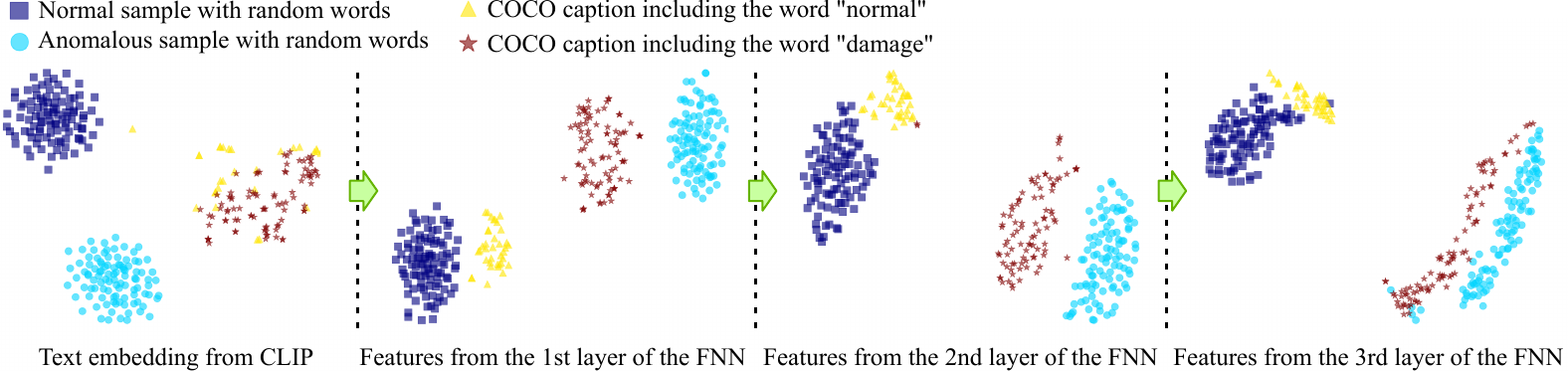}
  \caption{t-SNE plots of feature embeddings from different layers.}
  \label{fig:feature}
  \vspace{-1em}
\end{figure}

To analyze how the FNN learns to classify normal and anomalous samples with our random word data augmentation, we obtain natural sentences that contain a word of either ``normal'' or ``damage'' from COCO captions~\cite{lin_eccv2014} and plot the embeddings of those natural sentences in addition to the augmented prompts with t-SNE~\cite{maaten_jmlr2008}. Figure~\ref{fig:feature} shows the plotted results. As shown in the figure, CLIP's embeddings of the natural sentences and augmented prompts have different domains. However, the embeddings from the FNN layers are gradually split by their sample types rather than their domains. These plots indicate that the trained FNN learns to extract features containing information on normality and thus successfully classifies anomalous samples of arbitrary object categories.

\subsection{Potential Application}

\begin{table}[tb]
  \centering
  \caption{Comparison of CLIP and our method on the SewerML dataset.}\label{tab:compsw}
  \footnotesize
  \begin{tabular}{@{}lcccc@{}}
    \toprule
    Method & AUROC & AUPR & $F_{1}$-max \\
    \midrule
    CLIP~\cite{radford_icml2021} & 79.1 & 77.7 & 75.2 \\
    [0.5ex]\hdashline\noalign{\vskip 0.5ex}
    Ours & \textbf{82.6$\pm$1.3} & \textbf{81.0$\pm$1.3} & \textbf{77.4$\pm$0.9} \\
    CLIP + Ours & 81.7$\pm$0.9 & 80.2$\pm$0.7 & 76.9$\pm$0.8 \\
    \bottomrule
  \end{tabular}
  \vspace{-1.3em}
\end{table}

To validate the effectiveness of our method in detecting ambiguous anomalous samples, we evaluate the performances of CLIP~\cite{radford_icml2021} and our method on the SewerML dataset~\cite{haurum_cvpr2021}. Table~\ref{tab:compsw} shows the comparison results. As shown in the table, our method without the prompt-guided AD achieves the best performance among the three methods. The reason for the best performance without the prompt-guided AD is probably because CLIP cannot handle the diversity of defects in the dataset and thus degrades the performance rather than improving it. In contrast, our FNN is trained with a diverse set of normal and anomalous samples and thus can achieve better performance.

\section{Conclusion}\label{sec:conc}

We propose a novel zero-shot category-agnostic AD method that leverages CLIP as a data source for training anomaly detectors. Our method differs from existing methods in that our method does not include object categories in prompts to generate training samples and thus can be applied to the case where object categories are unknown. Furthermore, training samples generated by our method are so diverse that our method can achieve significant performance even without prompt ensembling. We perform extensive experiments and show the effectiveness of our method.

\bibliography{egbib}

\newpage

\setcounter{section}{0}
\renewcommand{\thesection}{\Alph{section}}

\section{Detailed Quantitative Results}

We report category-wise performance of CLIP~\cite{radford_icml2021}, WinCLIP~\cite{jeong_arxiv2023}, and ours for future reference. Our performances are those with the score combination of CLIP and our trained feed-forward network. The reported values of our method are the mean and standard deviation of trials with 10 random seeds.

\begin{table}[tbh]
  \centering
  \caption{Class-wise AUROC on the MVTec-AD dataset.}\label{tab:auroc_mvtec}
  \scriptsize
  \setlength{\tabcolsep}{1.4pt}
  \begin{tabular}{@{}lcccccccccc@{}}
    \toprule
     & \multicolumn{2}{c}{\multirow{2}{*}{\shortstack{0-shot $\left(\begin{array}{@{}c@{}}\text{Object}\\\text{unknown}\end{array}\right)$}}} & \multicolumn{2}{c}{\multirow{2}{*}{\shortstack{0-shot $\left(\begin{array}{@{}c@{}}\text{Object}\\\text{known}\end{array}\right)$}}} & & & & & & \\
     & \multicolumn{2}{c}{} & \multicolumn{2}{c}{} & \multicolumn{2}{c}{1-shot} & \multicolumn{2}{c}{2-shot} & \multicolumn{2}{c}{4-shot} \\
     \cmidrule(lr){2-3}\cmidrule(lr){4-5}\cmidrule(lr){6-7}\cmidrule(lr){8-9}\cmidrule(lr){10-11}
     Category & CLIP & CLIP + ours & WinCLIP & CLIP + ours & WinCLIP & CLIP + ours & WinCLIP & CLIP + ours & WinCLIP & CLIP + ours \\
     \midrule
     Bottle & 99.4 & 97.1$\pm$1.1 & 99.2 & 97.0$\pm$0.8 & 98.2$\pm$0.9 & 97.8$\pm$0.6 & 99.3$\pm$0.3 & 98.1$\pm$0.7 & 99.3$\pm$0.4 & 98.2$\pm$0.6 \\
     Cable & 83.0 & 84.2$\pm$0.6 & 86.5 & 86.4$\pm$1.4 & 88.9$\pm$1.9 & 90.8$\pm$1.1 & 88.4$\pm$0.7 & 91.3$\pm$2.0 & 90.9$\pm$0.9 & 92.1$\pm$0.8 \\
     Capsule & 85.3 & 87.3$\pm$2.0 & 72.9 & 88.7$\pm$2.8 & 72.3$\pm$6.8 & 76.8$\pm$6.4 & 77.3$\pm$8.8 & 83.6$\pm$7.1 & 82.3$\pm$8.9 & 87.8$\pm$7.9 \\
     Carpet & 100 & 100$\pm$0.0 & 100 & 100$\pm$0.0 & 99.8$\pm$0.3 & 100$\pm$0.0 & 99.8$\pm$0.3 & 100$\pm$0.0 & 100$\pm$0.0 & 100$\pm$0.0 \\
     Grid & 99.2 & 99.0$\pm$0.4 & 98.8 & 98.7$\pm$0.2 & 99.5$\pm$0.3 & 99.2$\pm$0.5 & 99.4$\pm$0.2 & 99.1$\pm$0.4 & 99.6$\pm$0.1 & 99.6$\pm$0.2 \\
     Hazelnut & 92.0 & 94.3$\pm$1.1 & 93.9 & 94.0$\pm$0.5 & 97.5$\pm$1.4 & 94.9$\pm$0.6 & 98.3$\pm$0.7 & 95.0$\pm$0.5 & 98.4$\pm$0.4 & 95.1$\pm$0.5 \\
     Leather & 100 & 100$\pm$0.0 & 100 & 100$\pm$0.0 & 99.9$\pm$0.0 & 100$\pm$0.0 & 99.9$\pm$0.0 & 99.9$\pm$0.0 & 100$\pm$0.0 & 100$\pm$0.0 \\
     Metal nut & 94.4 & 96.1$\pm$1.3 & 97.1 & 94.7$\pm$1.7 & 98.7$\pm$0.8 & 96.1$\pm$1.3 & 99.4$\pm$0.2 & 96.1$\pm$1.4 & 99.5$\pm$0.2 & 96.3$\pm$1.3 \\
     Pill & 88.3 & 88.6$\pm$1.6 & 79.1 & 90.2$\pm$1.0 & 91.2$\pm$2.1 & 92.5$\pm$1.3 & 92.3$\pm$0.7 & 92.6$\pm$1.2 & 92.8$\pm$1.0 & 92.6$\pm$1.3 \\
     Screw & 76.1 & 76.6$\pm$1.5 & 83.3 & 75.5$\pm$2.1 & 86.4$\pm$0.9 & 77.4$\pm$1.8 & 86.0$\pm$2.1 & 77.5$\pm$2.1 & 87.9$\pm$1.2 & 77.6$\pm$2.1 \\
     Tile & 99.4 & 99.5$\pm$0.2 & 100 & 99.5$\pm$0.2 & 99.9$\pm$0.0 & 99.6$\pm$0.1 & 99.9$\pm$0.2 & 99.6$\pm$0.1 & 99.9$\pm$0.1 & 99.6$\pm$0.1 \\
     Toothbrush & 92.8 & 88.4$\pm$3.5 & 87.5 & 94.0$\pm$0.9 & 92.2$\pm$4.9 & 94.7$\pm$1.3 & 97.5$\pm$1.6 & 95.0$\pm$1.0 & 96.7$\pm$2.6 & 96.4$\pm$1.5 \\
     Transistor & 79.7 & 86.6$\pm$1.4 & 88.0 & 88.9$\pm$1.3 & 83.4$\pm$3.8 & 89.4$\pm$1.3 & 85.3$\pm$1.7 & 89.6$\pm$1.4 & 85.7$\pm$2.5 & 89.6$\pm$1.3 \\
     Wood & 97.8 & 97.6$\pm$0.6 & 99.4 & 98.1$\pm$0.5 & 99.9$\pm$0.1 & 98.7$\pm$0.6 & 99.9$\pm$0.1 & 98.8$\pm$0.5 & 99.8$\pm$0.3 & 98.8$\pm$0.6 \\
     Zipper & 85.5 & 87.4$\pm$1.0 & 91.5 & 88.8$\pm$0.1 & 88.8$\pm$5.9 & 91.6$\pm$2.9 & 94.0$\pm$1.4 & 93.8$\pm$1.0 & 94.5$\pm$0.5 & 94.3$\pm$1.0 \\
     \bottomrule
  \end{tabular}
\end{table}

\begin{table}[tbh]
  \centering
  \caption{Class-wise AUPR on the MVTec-AD dataset.}\label{tab:auroc_mvtec}
  \scriptsize
  \setlength{\tabcolsep}{1.4pt}
  \begin{tabular}{@{}lcccccccccc@{}}
    \toprule
     & \multicolumn{2}{c}{\multirow{2}{*}{\shortstack{0-shot $\left(\begin{array}{@{}c@{}}\text{Object}\\\text{unknown}\end{array}\right)$}}} & \multicolumn{2}{c}{\multirow{2}{*}{\shortstack{0-shot $\left(\begin{array}{@{}c@{}}\text{Object}\\\text{known}\end{array}\right)$}}} & & & & & & \\
     & \multicolumn{2}{c}{} & \multicolumn{2}{c}{} & \multicolumn{2}{c}{1-shot} & \multicolumn{2}{c}{2-shot} & \multicolumn{2}{c}{4-shot} \\
     \cmidrule(lr){2-3}\cmidrule(lr){4-5}\cmidrule(lr){6-7}\cmidrule(lr){8-9}\cmidrule(lr){10-11}
     Category & CLIP & CLIP + ours & WinCLIP & CLIP + ours & WinCLIP & CLIP + ours & WinCLIP & CLIP + ours & WinCLIP & CLIP + ours \\
     \midrule
     Bottle & 99.8 & 99.2$\pm$0.3 & 99.8 & 99.2$\pm$0.2 & 99.4$\pm$0.3 & 99.4$\pm$0.2 & 99.8$\pm$0.1 & 99.4$\pm$0.2 & 99.8$\pm$0.1 & 99.5$\pm$0.2 \\
     Cable & 88.3 & 89.3$\pm$0.5 & 91.2 & 91.6$\pm$1.0 & 93.2$\pm$1.1 & 94.6$\pm$0.8 & 92.9$\pm$0.6 & 94.9$\pm$1.2 & 94.4$\pm$0.3 & 95.4$\pm$0.5 \\
     Capsule & 96.8 & 96.8$\pm$0.7 & 91.5 & 97.0$\pm$0.9 & 91.6$\pm$2.7 & 93.4$\pm$2.2 & 93.3$\pm$3.6 & 95.6$\pm$2.2 & 95.1$\pm$3.3 & 96.6$\pm$2.6 \\
     Carpet & 100 & 100$\pm$0.0 & 100 & 100$\pm$0.0 & 99.9$\pm$0.1 & 100$\pm$0.0 & 99.9$\pm$0.1 & 100$\pm$0.0 & 100$\pm$0.0 & 100$\pm$0.0 \\
     Grid & 99.7 & 99.7$\pm$0.1 & 99.6 & 99.6$\pm$0.0 & 99.9$\pm$0.1 & 99.7$\pm$0.2 & 99.8$\pm$0.1 & 99.7$\pm$0.1 & 99.9$\pm$0.0 & 99.9$\pm$0.1 \\
     Hazelnut & 96.0 & 96.7$\pm$0.6 & 96.9 & 96.2$\pm$0.4 & 98.6$\pm$0.7 & 96.8$\pm$0.5 & 99.1$\pm$0.4 & 96.8$\pm$0.4 & 99.1$\pm$0.2 & 96.9$\pm$0.5 \\
     Leather & 100 & 100$\pm$0.0 & 100 & 100$\pm$0.0 & 100$\pm$0.0 & 100$\pm$0.0 & 100$\pm$0.0 & 100$\pm$0.0 & 100$\pm$0.0 & 100$\pm$0.0 \\
     Metal nut & 98.8 & 99.1$\pm$0.3 & 99.3 & 98.8$\pm$0.4 & 99.7$\pm$0.2 & 99.1$\pm$0.3 & 99.9$\pm$0.0 & 99.1$\pm$0.3 & 99.9$\pm$0.1 & 99.2$\pm$0.3 \\
     Pill & 97.5 & 97.6$\pm$0.4 & 95.7 & 98.0$\pm$0.2 & 98.3$\pm$0.5 & 98.5$\pm$0.3 & 98.6$\pm$0.1 & 98.5$\pm$0.3 & 98.6$\pm$0.2 & 98.5$\pm$0.3 \\
     Screw & 90.3 & 90.7$\pm$0.8 & 93.1 & 90.1$\pm$1.2 & 94.2$\pm$0.6 & 91.0$\pm$1.1 & 94.1$\pm$1.5 & 91.1$\pm$1.1 & 94.9$\pm$0.8 & 91.1$\pm$1.1 \\
     Tile & 99.8 & 99.8$\pm$0.1 & 100 & 99.8$\pm$0.1 & 100$\pm$0.0 & 99.8$\pm$0.1 & 100$\pm$0.1 & 99.8$\pm$0.0 & 100$\pm$0.0 & 99.8$\pm$0.1 \\
     Toothbrush & 97.7 & 95.4$\pm$1.7 & 95.6 & 97.6$\pm$0.4 & 96.7$\pm$2.0 & 97.9$\pm$0.6 & 99.0$\pm$0.6 & 98.0$\pm$0.4 & 98.7$\pm$1.1 & 98.6$\pm$0.6 \\
     Transistor & 76.5 & 80.4$\pm$1.7 & 87.1 & 82.3$\pm$2.1 & 79.0$\pm$4.0 & 83.3$\pm$2.1 & 80.7$\pm$2.3 & 83.5$\pm$2.2 & 80.7$\pm$3.2 & 83.6$\pm$2.1 \\
     Wood & 99.3 & 99.3$\pm$0.2 & 99.8 & 99.4$\pm$0.2 & 100$\pm$0.0 & 99.6$\pm$0.2 & 100$\pm$0.0 & 99.6$\pm$0.2 & 99.9$\pm$0.1 & 99.6$\pm$0.2 \\
     Zipper & 95.6 & 96.1$\pm$0.3 & 97.5 & 96.7$\pm$0.0 & 96.8$\pm$1.8 & 97.5$\pm$0.9 & 98.3$\pm$0.4 & 98.1$\pm$0.4 & 98.5$\pm$0.2 & 98.3$\pm$0.3 \\
     \bottomrule
  \end{tabular}
\end{table}

\begin{table}[tbh]
  \centering
  \caption{Class-wise $F_{1}$-max on the MVTec-AD dataset.}\label{tab:auroc_mvtec}
  \scriptsize
  \setlength{\tabcolsep}{1.4pt}
  \begin{tabular}{@{}lcccccccccc@{}}
    \toprule
     & \multicolumn{2}{c}{\multirow{2}{*}{\shortstack{0-shot $\left(\begin{array}{@{}c@{}}\text{Object}\\\text{unknown}\end{array}\right)$}}} & \multicolumn{2}{c}{\multirow{2}{*}{\shortstack{0-shot $\left(\begin{array}{@{}c@{}}\text{Object}\\\text{known}\end{array}\right)$}}} & & & & & & \\
     & \multicolumn{2}{c}{} & \multicolumn{2}{c}{} & \multicolumn{2}{c}{1-shot} & \multicolumn{2}{c}{2-shot} & \multicolumn{2}{c}{4-shot} \\
     \cmidrule(lr){2-3}\cmidrule(lr){4-5}\cmidrule(lr){6-7}\cmidrule(lr){8-9}\cmidrule(lr){10-11}
     Category & CLIP & CLIP + ours & WinCLIP & CLIP + ours & WinCLIP & CLIP + ours & WinCLIP & CLIP + ours & WinCLIP & CLIP + ours \\
     \midrule
     Bottle & 98.4 & 95.2$\pm$1.1 & 97.6 & 95.9$\pm$0.7 & 96.5$\pm$1.3 & 96.5$\pm$0.6 & 97.7$\pm$0.7 & 96.6$\pm$0.7 & 97.8$\pm$0.6 & 96.6$\pm$0.7 \\
     Cable & 84.5 & 85.7$\pm$0.8 & 84.5 & 84.4$\pm$1.2 & 86.1$\pm$1.3 & 87.7$\pm$1.0 & 85.2$\pm$0.7 & 87.6$\pm$2.0 & 87.2$\pm$0.6 & 87.8$\pm$1.0 \\
     Capsule & 90.7 & 92.6$\pm$0.8 & 91.4 & 93.3$\pm$0.8 & 91.6$\pm$0.7 & 93.1$\pm$0.7 & 92.1$\pm$0.7 & 94.0$\pm$0.8 & 92.5$\pm$0.5 & 94.7$\pm$0.8 \\
     Carpet & 99.4 & 99.6$\pm$0.4 & 99.4 & 99.6$\pm$0.4 & 99.2$\pm$0.8 & 100$\pm$0.0 & 99.3$\pm$0.7 & 100$\pm$0.0 & 99.9$\pm$0.2 & 100$\pm$0.0 \\
     Grid & 97.3 & 97.8$\pm$0.6 & 98.2 & 97.3$\pm$0.0 & 98.9$\pm$0.4 & 98.1$\pm$0.7 & 99.1$\pm$0.0 & 98.1$\pm$0.5 & 99.1$\pm$0.0 & 98.4$\pm$0.4 \\
     Hazelnut & 88.9 & 91.9$\pm$1.7 & 89.7 & 91.8$\pm$1.0 & 94.7$\pm$2.3 & 92.7$\pm$0.9 & 95.6$\pm$1.6 & 92.9$\pm$0.8 & 96.2$\pm$1.0 & 93.0$\pm$0.8 \\
     Leather & 100 & 100$\pm$0.0 & 100 & 100$\pm$0.0 & 99.5$\pm$0.0 & 99.7$\pm$0.3 & 99.7$\pm$0.2 & 99.5$\pm$0.3 & 99.8$\pm$0.2 & 99.9$\pm$0.2 \\
     Metal nut & 93.3 & 95.0$\pm$1.1 & 96.3 & 94.0$\pm$1.2 & 97.7$\pm$1.0 & 95.0$\pm$1.0 & 98.4$\pm$0.5 & 94.9$\pm$1.2 & 98.5$\pm$0.6 & 95.1$\pm$1.0 \\
     Pill & 93.7 & 93.4$\pm$0.6 & 91.6 & 93.9$\pm$0.3 & 93.8$\pm$0.7 & 95.1$\pm$0.7 & 94.3$\pm$0.4 & 95.2$\pm$0.7 & 94.1$\pm$0.4 & 95.1$\pm$0.8 \\
     Screw & 86.9 & 86.9$\pm$0.0 & 87.4 & 86.9$\pm$0.2 & 88.5$\pm$0.3 & 86.9$\pm$0.3 & 89.0$\pm$0.6 & 87.0$\pm$0.4 & 89.6$\pm$0.7 & 87.0$\pm$0.4 \\
     Tile & 97.7 & 97.8$\pm$0.6 & 99.4 & 97.8$\pm$0.6 & 98.9$\pm$0.2 & 97.9$\pm$0.5 & 99.2$\pm$0.3 & 97.9$\pm$0.5 & 99.2$\pm$0.3 & 98.0$\pm$0.5 \\
     Toothbrush & 93.1 & 90.2$\pm$0.9 & 87.9 & 92.7$\pm$0.7 & 94.1$\pm$1.9 & 94.1$\pm$1.1 & 96.7$\pm$1.8 & 93.7$\pm$0.9 & 96.8$\pm$2.3 & 95.3$\pm$1.6 \\
     Transistor & 69.7 & 79.6$\pm$1.8 & 79.5 & 81.7$\pm$1.5 & 75.1$\pm$3.1 & 81.6$\pm$1.7 & 75.9$\pm$2.4 & 81.7$\pm$1.7 & 76.6$\pm$2.8 & 81.8$\pm$1.5 \\
     Wood & 95.7 & 95.8$\pm$0.6 & 98.3 & 96.0$\pm$0.4 & 99.4$\pm$0.3 & 97.2$\pm$0.6 & 99.5$\pm$0.4 & 97.2$\pm$0.6 & 99.2$\pm$0.9 & 97.1$\pm$0.7 \\
     Zipper & 90.0 & 91.2$\pm$0.8 & 92.9 & 91.9$\pm$0.1 & 92.1$\pm$2.5 & 93.7$\pm$2.1 & 94.4$\pm$0.3 & 95.5$\pm$0.3 & 94.7$\pm$0.4 & 95.5$\pm$0.3 \\
     \bottomrule
  \end{tabular}
\end{table}

\begin{table}[tbh]
  \centering
  \caption{Class-wise AUROC on the VisA dataset.}\label{tab:auroc_mvtec}
  \scriptsize
  \setlength{\tabcolsep}{1.4pt}
  \begin{tabular}{@{}lcccccccccc@{}}
    \toprule
     & \multicolumn{2}{c}{\multirow{2}{*}{\shortstack{0-shot $\left(\begin{array}{@{}c@{}}\text{Object}\\\text{unknown}\end{array}\right)$}}} & \multicolumn{2}{c}{\multirow{2}{*}{\shortstack{0-shot $\left(\begin{array}{@{}c@{}}\text{Object}\\\text{known}\end{array}\right)$}}} & & & & & & \\
     & \multicolumn{2}{c}{} & \multicolumn{2}{c}{} & \multicolumn{2}{c}{1-shot} & \multicolumn{2}{c}{2-shot} & \multicolumn{2}{c}{4-shot} \\
     \cmidrule(lr){2-3}\cmidrule(lr){4-5}\cmidrule(lr){6-7}\cmidrule(lr){8-9}\cmidrule(lr){10-11}
     Category & CLIP & CLIP + ours & WinCLIP & CLIP + ours & WinCLIP & CLIP + ours & WinCLIP & CLIP + ours & WinCLIP & CLIP + ours \\
     \midrule
     Candle & 96.0 & 96.9$\pm$0.3 & 95.4 & 97.0$\pm$0.1 & 93.4$\pm$1.4 & 97.7$\pm$0.2 & 94.8$\pm$1.0 & 97.9$\pm$0.1 & 95.1$\pm$0.3 & 97.9$\pm$0.2 \\
     Capsules & 74.9 & 78.1$\pm$2.3 & 85.0 & 75.6$\pm$0.3 & 85.0$\pm$3.1 & 82.1$\pm$0.6 & 84.9$\pm$0.8 & 82.5$\pm$0.6 & 86.8$\pm$1.7 & 82.6$\pm$0.6 \\
     Cashew & 87.3 & 91.6$\pm$2.4 & 92.1 & 91.8$\pm$1.0 & 94.0$\pm$0.4 & 92.9$\pm$1.1 & 94.3$\pm$0.5 & 93.1$\pm$0.8 & 95.2$\pm$0.8 & 93.1$\pm$0.8 \\
     Chewing gum & 89.8 & 94.3$\pm$1.3 & 96.5 & 93.0$\pm$1.9 & 97.6$\pm$0.8 & 95.0$\pm$0.7 & 97.3$\pm$0.8 & 95.1$\pm$0.7 & 97.7$\pm$0.3 & 95.3$\pm$0.6 \\
     Fryum & 88.2 & 90.0$\pm$0.5 & 80.3 & 89.2$\pm$0.5 & 88.5$\pm$1.9 & 92.6$\pm$1.2 & 90.5$\pm$0.4 & 92.8$\pm$1.3 & 90.8$\pm$0.5 & 92.9$\pm$1.2 \\
     Macaroni1 & 82.0 & 82.5$\pm$2.0 & 76.2 & 89.4$\pm$1.4 & 82.9$\pm$1.5 & 91.5$\pm$1.7 & 83.3$\pm$1.9 & 91.5$\pm$1.7 & 85.2$\pm$0.9 & 91.6$\pm$1.6 \\
     Macaroni2 & 65.6 & 71.3$\pm$1.5 & 63.7 & 68.6$\pm$1.5 & 70.2$\pm$0.9 & 69.8$\pm$1.5 & 71.8$\pm$2.0 & 69.9$\pm$1.4 & 70.9$\pm$2.2 & 69.9$\pm$1.3 \\
     PCB1 & 58.2 & 49.7$\pm$7.0 & 73.6 & 69.4$\pm$3.8 & 75.6$\pm$23.0 & 70.7$\pm$16.9 & 76.7$\pm$5.2 & 80.2$\pm$10.4 & 88.3$\pm$1.7 & 83.1$\pm$2.5 \\
     PCB2 & 51.6 & 50.7$\pm$1.0 & 51.2 & 47.3$\pm$0.1 & 62.2$\pm$3.9 & 60.7$\pm$2.2 & 62.6$\pm$3.7 & 63.8$\pm$1.9 & 67.5$\pm$2.6 & 66.4$\pm$1.5 \\
     PCB3 & 66.0 & 67.9$\pm$1.3 & 73.4 & 66.3$\pm$0.4 & 74.1$\pm$1.1 & 74.0$\pm$8.4 & 78.8$\pm$1.9 & 80.3$\pm$3.4 & 83.3$\pm$1.7 & 83.4$\pm$1.8 \\
     PCB4 & 74.5 & 75.2$\pm$0.7 & 79.6 & 82.9$\pm$0.4 & 85.2$\pm$8.9 & 88.1$\pm$9.0 & 82.3$\pm$9.9 & 87.6$\pm$10.4 & 87.6$\pm$8.0 & 89.7$\pm$10.1 \\
     Pipe fryum & 84.1 & 89.9$\pm$1.4 & 69.7 & 87.2$\pm$2.6 & 97.2$\pm$1.1 & 92.6$\pm$1.7 & 98.0$\pm$0.6 & 93.0$\pm$1.8 & 98.5$\pm$0.4 & 93.2$\pm$1.8 \\
     \bottomrule
  \end{tabular}
\end{table}

\begin{table}[tbh]
  \centering
  \caption{Class-wise AUPR on the VisA dataset.}\label{tab:auroc_mvtec}
  \scriptsize
  \setlength{\tabcolsep}{1.4pt}
  \begin{tabular}{@{}lcccccccccc@{}}
    \toprule
     & \multicolumn{2}{c}{\multirow{2}{*}{\shortstack{0-shot $\left(\begin{array}{@{}c@{}}\text{Object}\\\text{unknown}\end{array}\right)$}}} & \multicolumn{2}{c}{\multirow{2}{*}{\shortstack{0-shot $\left(\begin{array}{@{}c@{}}\text{Object}\\\text{known}\end{array}\right)$}}} & & & & & & \\
     & \multicolumn{2}{c}{} & \multicolumn{2}{c}{} & \multicolumn{2}{c}{1-shot} & \multicolumn{2}{c}{2-shot} & \multicolumn{2}{c}{4-shot} \\
     \cmidrule(lr){2-3}\cmidrule(lr){4-5}\cmidrule(lr){6-7}\cmidrule(lr){8-9}\cmidrule(lr){10-11}
     Category & CLIP & CLIP + ours & WinCLIP & CLIP + ours & WinCLIP & CLIP + ours & WinCLIP & CLIP + ours & WinCLIP & CLIP + ours \\
     \midrule
     Candle & 96.3 & 96.9$\pm$0.3 & 95.8 & 96.9$\pm$0.2 & 93.6$\pm$1.5 & 97.6$\pm$0.2 & 95.1$\pm$1.1 & 97.7$\pm$0.1 & 95.3$\pm$0.4 & 97.8$\pm$0.1 \\
     Capsules & 83.1 & 85.4$\pm$1.5 & 90.9 & 84.8$\pm$0.1 & 89.9$\pm$2.5 & 88.6$\pm$0.4 & 88.9$\pm$0.7 & 88.9$\pm$0.3 & 91.5$\pm$1.4 & 89.0$\pm$0.3 \\
     Cashew & 94.3 & 96.2$\pm$1.0 & 96.4 & 96.3$\pm$0.4 & 97.2$\pm$0.2 & 96.9$\pm$0.4 & 97.3$\pm$0.2 & 96.9$\pm$0.3 & 97.7$\pm$0.4 & 96.9$\pm$0.3 \\
     Chewing gum & 95.6 & 97.7$\pm$0.6 & 98.6 & 97.0$\pm$0.9 & 99.0$\pm$0.3 & 97.9$\pm$0.3 & 98.9$\pm$0.3 & 98.0$\pm$0.3 & 99.0$\pm$0.1 & 98.0$\pm$0.2 \\
     Fryum & 94.8 & 95.7$\pm$0.2 & 90.1 & 95.4$\pm$0.3 & 94.7$\pm$1.0 & 97.0$\pm$0.5 & 95.8$\pm$0.2 & 97.0$\pm$0.6 & 96.0$\pm$0.3 & 97.1$\pm$0.6 \\
     Macaroni1 & 85.0 & 84.5$\pm$1.7 & 75.8 & 90.1$\pm$1.1 & 84.9$\pm$1.2 & 92.0$\pm$1.4 & 84.7$\pm$1.5 & 92.0$\pm$1.4 & 86.5$\pm$0.6 & 92.1$\pm$1.3 \\
     Macaroni2 & 60.8 & 67.4$\pm$1.6 & 60.3 & 66.3$\pm$1.9 & 68.4$\pm$1.8 & 68.5$\pm$1.6 & 70.4$\pm$1.8 & 68.9$\pm$1.7 & 69.6$\pm$2.8 & 68.9$\pm$1.7 \\
     PCB1 & 63.6 & 55.3$\pm$6.3 & 78.4 & 71.9$\pm$4.0 & 76.5$\pm$19.0 & 72.4$\pm$13.8 & 78.3$\pm$4.3 & 80.0$\pm$8.8 & 87.7$\pm$1.7 & 82.4$\pm$2.6 \\
     PCB2 & 54.9 & 54.0$\pm$0.7 & 49.2 & 49.6$\pm$0.1 & 64.9$\pm$3.3 & 64.7$\pm$2.1 & 65.8$\pm$4.0 & 68.0$\pm$1.5 & 71.3$\pm$3.4 & 70.1$\pm$0.7 \\
     PCB3 & 66.4 & 69.6$\pm$1.3 & 76.5 & 67.3$\pm$0.3 & 73.5$\pm$1.6 & 74.6$\pm$8.7 & 80.9$\pm$1.6 & 82.5$\pm$3.0 & 84.8$\pm$1.8 & 85.4$\pm$1.4 \\
     PCB4 & 79.6 & 79.9$\pm$0.5 & 77.7 & 84.4$\pm$0.4 & 78.5$\pm$15.5 & 83.2$\pm$13.7 & 72.5$\pm$16.2 & 83.6$\pm$14.4 & 85.6$\pm$8.9 & 86.0$\pm$14.6 \\
     Pipe fryum & 91.6 & 95.1$\pm$0.8 & 82.3 & 93.6$\pm$2.1 & 98.6$\pm$0.5 & 96.2$\pm$1.1 & 99.0$\pm$0.3 & 96.4$\pm$1.1 & 99.2$\pm$0.2 & 96.5$\pm$1.1 \\
     \bottomrule
  \end{tabular}
\end{table}

\begin{table}[tbh]
  \centering
  \caption{Class-wise $F_{1}$-max on the VisA dataset.}\label{tab:auroc_mvtec}
  \scriptsize
  \setlength{\tabcolsep}{1.4pt}
  \begin{tabular}{@{}lcccccccccc@{}}
    \toprule
     & \multicolumn{2}{c}{\multirow{2}{*}{\shortstack{0-shot $\left(\begin{array}{@{}c@{}}\text{Object}\\\text{unknown}\end{array}\right)$}}} & \multicolumn{2}{c}{\multirow{2}{*}{\shortstack{0-shot $\left(\begin{array}{@{}c@{}}\text{Object}\\\text{known}\end{array}\right)$}}} & & & & & & \\
     & \multicolumn{2}{c}{} & \multicolumn{2}{c}{} & \multicolumn{2}{c}{1-shot} & \multicolumn{2}{c}{2-shot} & \multicolumn{2}{c}{4-shot} \\
     \cmidrule(lr){2-3}\cmidrule(lr){4-5}\cmidrule(lr){6-7}\cmidrule(lr){8-9}\cmidrule(lr){10-11}
     Category & CLIP & CLIP + ours & WinCLIP & CLIP + ours & WinCLIP & CLIP + ours & WinCLIP & CLIP + ours & WinCLIP & CLIP + ours \\
     \midrule
     Candle & 91.1 & 92.4$\pm$0.6 & 89.4 & 92.2$\pm$0.3 & 87.8$\pm$1.2 & 94.7$\pm$0.5 & 89.1$\pm$1.3 & 95.2$\pm$0.4 & 88.9$\pm$1.0 & 95.1$\pm$0.3 \\
     Capsules & 79.8 & 81.4$\pm$1.3 & 83.9 & 79.3$\pm$0.2 & 84.9$\pm$2.0 & 82.8$\pm$0.6 & 85.4$\pm$0.6 & 83.0$\pm$0.6 & 86.0$\pm$0.9 & 83.1$\pm$0.5 \\
     Cashew & 84.3 & 89.0$\pm$2.3 & 88.4 & 89.1$\pm$1.2 & 90.7$\pm$0.7 & 89.6$\pm$1.3 & 90.9$\pm$0.7 & 89.8$\pm$1.0 & 91.6$\pm$1.3 & 90.0$\pm$1.0 \\
     Chewing gum & 87.4 & 92.9$\pm$1.7 & 94.8 & 90.2$\pm$2.4 & 95.6$\pm$0.9 & 92.7$\pm$1.0 & 95.4$\pm$0.6 & 93.0$\pm$0.9 & 95.7$\pm$0.5 & 93.4$\pm$0.6 \\
     Fryum & 86.2 & 87.6$\pm$0.8 & 82.7 & 86.9$\pm$0.9 & 87.2$\pm$1.4 & 91.0$\pm$2.0 & 88.4$\pm$0.6 & 91.0$\pm$1.9 & 88.9$\pm$0.8 & 91.4$\pm$1.9 \\
     Macaroni1 & 74.4 & 76.7$\pm$1.1 & 74.2 & 83.1$\pm$1.2 & 76.2$\pm$1.4 & 84.9$\pm$1.5 & 76.7$\pm$2.0 & 84.8$\pm$1.5 & 78.2$\pm$1.2 & 84.9$\pm$1.6 \\
     Macaroni2 & 71.9 & 71.6$\pm$1.1 & 69.8 & 69.4$\pm$0.7 & 72.3$\pm$1.1 & 70.1$\pm$0.9 & 73.9$\pm$0.9 & 69.9$\pm$0.7 & 73.1$\pm$1.6 & 70.0$\pm$0.7 \\
     PCB1 & 66.7 & 66.7$\pm$0.0 & 71.0 & 69.2$\pm$1.1 & 81.3$\pm$6.6 & 75.6$\pm$6.9 & 73.2$\pm$3.7 & 78.7$\pm$4.8 & 83.1$\pm$2.2 & 78.5$\pm$2.3 \\
     PCB2 & 67.1 & 67.2$\pm$0.2 & 67.1 & 67.1$\pm$0.1 & 67.2$\pm$0.3 & 68.1$\pm$0.8 & 67.3$\pm$0.3 & 68.5$\pm$0.5 & 67.7$\pm$0.6 & 69.3$\pm$0.8 \\
     PCB3 & 70.5 & 69.9$\pm$0.9 & 71.0 & 68.9$\pm$0.3 & 73.5$\pm$1.5 & 72.7$\pm$3.4 & 73.9$\pm$1.3 & 75.3$\pm$2.0 & 77.0$\pm$1.4 & 76.9$\pm$1.8 \\
     PCB4 & 73.3 & 74.3$\pm$0.6 & 74.9 & 77.0$\pm$0.5 & 86.1$\pm$2.1 & 89.6$\pm$3.9 & 86.8$\pm$3.8 & 88.4$\pm$6.6 & 84.6$\pm$7.0 & 89.6$\pm$5.8 \\
     Pipe fryum & 85.1 & 88.6$\pm$1.6 & 80.7 & 86.9$\pm$1.3 & 94.4$\pm$0.7 & 90.9$\pm$1.8 & 95.4$\pm$0.8 & 91.3$\pm$1.6 & 95.6$\pm$0.7 & 91.5$\pm$1.4 \\
     \bottomrule
  \end{tabular}
\end{table}

\end{document}